\documentclass[11pt,a4paper]{article}
\usepackage[hyperref]{acl2021}
\usepackage{times}
\usepackage{latexsym}

\usepackage{microtype}
\usepackage{graphicx}
\usepackage{multirow}
\usepackage{amsmath}
\usepackage{float}
\usepackage{subfigure}
\usepackage{wrapfig}
\usepackage{CJKutf8}
\usepackage{color}
\usepackage{algorithm}
\usepackage{algorithmic}

\newcommand{\tabincell}[2]{\begin{tabular}{@{}#1@{}}#2\end{tabular}}

\aclfinalcopy 
\def\aclpaperid{1191} 

\title{Neural-Symbolic Solver for Math Word Problems with Auxiliary Tasks}

\author{
Jinghui Qin\textsuperscript{\rm 1}, Xiaodan Liang\textsuperscript{\rm 1,2}, Yining Hong\textsuperscript{\rm 3}, Jianheng Tang\textsuperscript{\rm 1} and Liang Lin\textsuperscript{\rm 1,2}\thanks{\ \ Corresponding Author}\\
\textsuperscript{\rm 1} Sun Yat-sen University \\
\textsuperscript{\rm 2} Dark Matter AI Inc.\\
\textsuperscript{\rm 3} University of California, Los Angeles\\
\texttt{qinjingh@mail2.sysu.edu.cn},\texttt{\{xdliang328,sqrt3tjh\}@gmail.com},\\\texttt{yininghong@cs.ucla.edu},\texttt{linliang@ieee.org}
}

\date{}

\begin{document}
\maketitle
\begin{abstract}
Previous math word problem solvers following the encoder-decoder paradigm fail to explicitly incorporate essential math symbolic constraints, leading to unexplainable and unreasonable predictions. Herein, we propose Neural-Symbolic Solver (NS-Solver) to explicitly and seamlessly incorporate different levels of symbolic constraints by auxiliary tasks. Our NS-Solver consists of a problem reader to encode problems, a programmer to generate symbolic equations, and a symbolic executor to obtain answers. Along with target expression supervision, our solver is also optimized via 4 new auxiliary objectives to enforce different symbolic reasoning: a) self-supervised number prediction task predicting both number quantity and number locations; b) commonsense constant prediction task predicting what prior knowledge (e.g. how many legs a chicken has) is required; c) program consistency checker computing the semantic loss between predicted equation and target equation to ensure reasonable equation mapping; d) duality exploiting task exploiting the quasi duality between symbolic equation generation and problem's part-of-speech generation to enhance the understanding ability of a solver. Besides, to provide a more realistic and challenging benchmark for developing a universal and scalable solver, we also construct a new large-scale MWP benchmark CM17K consisting of 4 kinds of MWPs (arithmetic, one-unknown linear, one-unknown non-linear, equation set) with more than 17K samples. Extensive experiments on Math23K and our CM17k demonstrate the superiority of our NS-Solver compared to state-of-the-art methods\footnote{The code and the new CM17k dataset are available at \url{https://github.com/QinJinghui/NS-Solver}.}.
\end{abstract}

\section{Introduction}
Deep neural networks have achieved remarkable successes in natural language processing recently. Although neural models have demonstrated performance superior to humans on some tasks, e.g. reading comprehension~\cite{squad,BERT,alBERT}, it still lacks the ability of discrete reasoning, resulting in low accuracy on math reasoning. Thus, it is hard for pure neural network approaches to tackle the task of solving math word problems (MWPs), which requires a model to be capable of natural language understanding and discrete reasoning. MWP solving aims to automatically answer a math word problem by understanding the textual description of the problem and reasoning out the underlying answer. A typical MWP is a short story that describes a partial state of the world and poses a question about an unknown quantity or multiple unknown quantities. To solve an MWP, the relevant quantities need to be identified from the text. Furthermore, the correct operators along with their computation order among these quantities need to be determined. Therefore, integrating neural networks with symbolic reasoning is crucial for solving MWPs. Inspired by the recent amazing progress on neural semantic parsing~\cite{liang-etal-2017-neural} and reading comprehension~\cite{chen2019neural}, we address this problem by neural-symbolic computing.  

Recently, many researchers~\cite{dns,cass,mathdqn,trnn,seq2tree,stackdecoder}, inspired by an encoder-decoder framework~\cite{seq2seq}, apply neural networks to solve MWPs by learning the mapping function between problems and their corresponding equations, and achieve remarkable successes. The encoder uses a neural network to represent a problem as a real-valued vector, and the decoder uses another neural network to generate an equation or expression token by token. The main difference among previous methods is the way to decode expressions or equations. However, they only follow the encoder-decoder paradigm but lacking the ability to explicitly incorporate essential math symbolic constraints (e.g. commonsense constants, formulation regularization), leading to unexplainable and unreasonable predictions. Besides, most of them only focus on arithmetic MWPs without any unknown, preventing them from generalizing to various types of MWPs, such as equation set problems. 

To address the above issues, we propose a novel \textbf{N}eural-\textbf{S}ymbolic \textbf{Solver} (NS-Solver), which explicitly and seamlessly incorporates different levels of symbolic constraints by auxiliary learning tasks. Our NS-Solver consists of three main components, a \textit{problem reader} to encode the math word problems into vector representations, a \textit{programmer} to generate the symbolic grounded equations, which are executed to produce answers, and a \textit{symbolic executor} to obtain final results. In addition to the supervised training objective between generated symbolic grounded equations and ground-truth equations, our solver is also optimized by four novel auxiliary objectives that enforce four levels of problem understanding and symbolic reasoning. First, we apply \textbf{number prediction task} to predict both the number quantity and number location in the problem in a self-supervised manner. Second, we deploy \textbf{commonsense constant prediction task} to predict what prior commonsense knowledge (e.g. how many legs a chicken has) is required for our solver. Third, we propose \textbf{program consistency checker} to compute the semantic loss between the predicted program and ground-truth equation to ensure reasonable equation mapping. Finally, we also propose a novel \textbf{duality exploiting task} that exploits the quasi duality between symbolic grounded equation generation and the problem's part-of-speech generation to enhance the understanding ability of our solver. There are some key advantages of our solution. First of all, the above four auxiliary tasks can produce additional training signals, which improves the data efficiency in training and makes our solver more robust. Second, using the predicted constant to constrain the target symbolic table can reduce the search space greatly, which means that our solver can generate correct symbolic grounded equations easier and better. Third, the auxiliary tasks have been proven to help reduce the domain gap between seen and unseen MWPs~\cite{sun2019unsupervised,sun2019test}, thus improving the reasoning ability of our solver. 

Besides, beyond the current large-scale high-quality MWP benchmark that only includes one type of problems, we also construct a large-scale challenging Chinese MWPs dataset CM17K, which contains 4 types of MWPs (arithmetic MWPs, one-unknown linear MWPs, one-unknown non-linear MWPs, equation set problems) with more than 17K samples, to provide a more realistic and challenging benchmark for developing a universal and scalable math solver. Extensive experiments on public Math23K and our proposed CM17k demonstrate the superiority of our NS-Solver compared to state-of-the-art methods in predicting final results while ensuring intermediate equation rationality.

\section{Related Work}
\noindent\textbf{Deep learning-based MWP Solvers.} Numerous methods have been proposed to tackle the MWP solving task, ranging from rule-based methods~\cite{bakman2007robust,yuhui2010frame-based}, statistical machine learning methods~\cite{kushman2014learning,zhou-etal-2015-learn,roy-roth-2015-solving,roy2016unit,mitra-baral-2016-learning,huang-etal-2016-well,roy2018mapping}, semantic parsing methods~\cite{shi-etal-2015-automatically,koncelkedziorski2015parsing,huang-etal-2017-learning,liang-etal-2018-meaning}, to deep learning methods~\cite{ling-etal-2017-program,dns,mathdqn,cass,seq2et,seq2tree,trnn,tsrmd,graph2tree,sau-solver,shen-jin-2020-solving,wu-etal-2020-knowledge,chen2021geoqa,hong2021lbf,hong2021smart}. However, most deep learning-based methods only follow the encoder-decoder framework without explicitly incorporating essential math symbolic constraints, resulting in some unexplainable and unreasonable predictions. Besides, most of them only focus on arithmetic MWPs, preventing them from generalizing to various types, such as equation set problems.
\begin{figure*}[t] 
	\centerline{\includegraphics[width=0.85\textwidth]{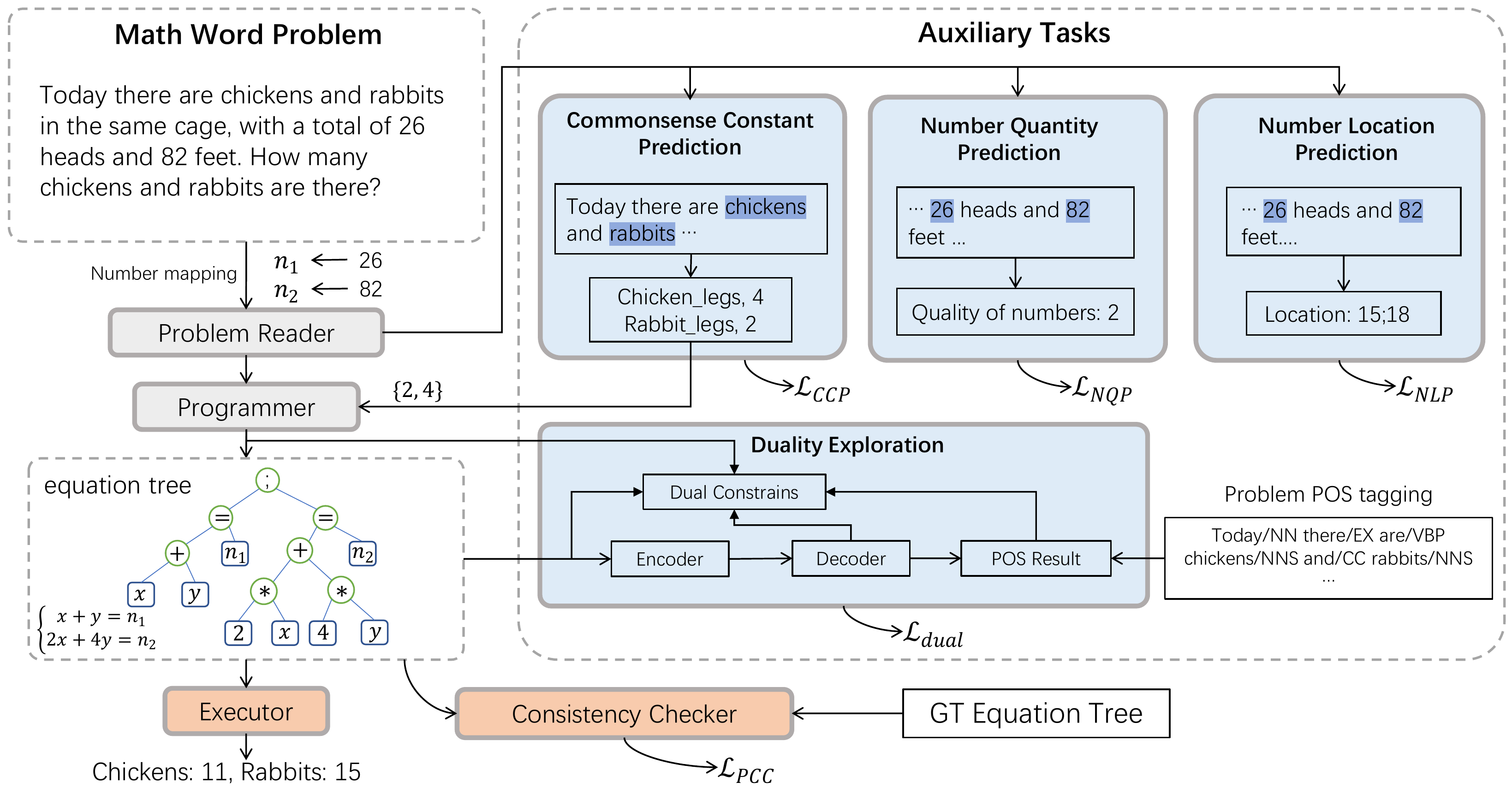}}
	\caption{An overview of our NS-Solver. When a problem preprocessed by number mapping and replacement is entered, our problem reader encodes the problem text into context representation. Then our programmer generates a tree-structured symbolic grounded program explicitly. Finally, a symbolic grounded program will be executed to produce answers by the executor. In our NS-Solver, we apply four auxiliary tasks to enhance its problem understanding and symbol reasoning ability for generating better programs. }
	\label{fig:sns-solver}
	\vspace{-3mm}
\end{figure*}

\noindent\textbf{Neural-Symbolic Computing.} Neural-symbolic computing has greatly promoted the development of semantic parsing. \citet{jia2016data,dong2016language,zhong2017seq2sql} applied neural sequence-to-sequence and sequence-to-tree models to semantic parsing with full supervision. \citet{liang2016neural,liang2018memory} have advanced the state-of-the-art in weakly supervised semantic parsing on knowledge graphs and tabular databases. Although most of the successes of semantic parsing are limited to structured data sources, it is not expensive for MWPs since it is easy to crawl lots of problems with annotated equations and answers.  Therefore, MWP solving can benefit from supervised neural-symbolic computing.

\noindent\textbf{Self-Supervised Learning.} Self-supervised auxiliary tasks have been widely used in the fields of natural language understanding~\cite{BERT,alBERT}. \citet{BERT} applied two self-supervised auxiliary tasks, masked LM and next sentence prediction, to improve the understanding ability of BERT by pretraining. ALBERT~\cite{alBERT} introduces sentence-order prediction task to address the ineffectiveness of the next sentence prediction task in BERT. \citet{hendrycks2019using} show that self-supervised learning can improve model robustness and uncertainty.

\noindent\textbf{Dual Learning.} Dual learning, first proposed by~\citet{he2016dual}, is a reinforcement training process that jointly trains a primal task and its dual task. Then ~\citet{xia2017dual} considered it as a way of supervised learning and designed a probabilistic regularization term to exploit the duality. It has been widely applied in various fields, such as machine translation~\cite{he2016dual}, sentiment classification~\cite{xia2017dual}, question answering~\cite{tang2017question}, visual question answering~\cite{li2018visual}, machine reading comprehension~\cite{xiao2018dual}, and code generation~\cite{wei2019code}. To the best of our knowledge, we are the first to exploit the duality in MWPs. Different from previous works, we design a quasi dual learning method between symbolic grounded equation generation and problem's part-of-speech generation to enhance the understanding ability by easing the difficulty of generating problems from symbolic equations.

\section{Neural-Symbolic Solver}
In this section, we present the design of the proposed NS-Solver. Its backbone mainly consists of a \textit{problem reader} that encodes the math word problems into vector representations, a \textit{programmer} to generate the symbolic grounded programs in prefix order, and a \textit{symbolic executor} to obtain final results. The overview of our NS-Solver is visualized in Fig.~\ref{fig:sns-solver}. We first introduce the backbone of our NS-Solver in section~\ref{sec:na}, and then we introduce other auxiliary tasks in section~\ref{sec:ssar}.
\subsection{Backbone}
\label{sec:na}
\noindent\textbf{Problem Reader.} Given a problem text $P$ = $\left \{ x_i \right \}^n_{i=1}$ processed by number template replacement which maps numeric values in a problem to number templates (e.g., $26$ and $82$ to $n_1$ and $n_2$ in Fig.~\ref{fig:sns-solver}), the problem reader encodes each token $x_i$ in the problem text into an embedding $e_i$. In this work, we deploy a two-layer bidirectional GRU to encode each token $x_i$ into an embedding $e_i=\overrightarrow{\mathbf{h}_{i}} + \overleftarrow{\mathbf{h}_{i}}$ where $\overrightarrow{\mathbf{h}_{i}}$  and $\overleftarrow{\mathbf{h}_{i}}$ are from forward and backward GRUs, respectively. Besides, our problem encoder also outputs a problem representation $\mathbf{g}_{0}= \overrightarrow{\mathbf{h}_{n}} + \overleftarrow{\mathbf{h}_{0}}$ as the initial hidden state of our programmer, where $\overrightarrow{\mathbf{h}_{n}}$ and $\overleftarrow{\mathbf{h}_{0}}$ are the last hidden state of forward and backward GRUs, respectively.

\noindent\textbf{Programmer.} The programmer takes the output of the problem reader as input and the problem representation as the initial hidden state, and then decodes a problem as a sequence of tokens $\left \{ y_i \right \}^{m}_{i=1}$ which are organized as a prefix equation tree. In this work, we deploy a tree-structured decoder~\cite{seq2tree} with attention mechanism~\cite{bahdanau2014neural} as the backbone of our programmer and modify them with UET representation~\cite{sau-solver} to support more symbols for multiple types of MWPs. In our programmer, the symbolic table consists of four parts. For each problem, the problem-specific symbolic table contains math operators ($ +,-,*,/,\hat ,= , ;$), unknown variable ($x$ and $y$), a series of commonsense constants ($1$, $3.14$, etc) predicted by the Commonsense Constant Prediction Task in \ref{sec:ssar}, and the problem-specific number templates ($n_1$, $n_2$, $n_3$, etc). It should be noticed that ; is a special operator with the lowest priority to integrate multiple equation trees as an ensemble equation tree, so that equation set problems can be handled as simple as arithmetic problems. 

\noindent\textbf{Executor.} We deploy sympy\footnote{https://www.sympy.org/}, which is a python library for symbolic mathematics, as our symbolic executor for obtaining final results by solving generated equations.

\subsection{The Design of Auxiliary Tasks}
\label{sec:ssar}
The MWP solving task remains challenging since previous methods did not take full advantage of the rich semantics contained in a problem and lacking the ability to explicitly incorporate essential math symbolic constraints. In this section, we introduce four auxiliary learning tasks to exploit additional training signals obtained from different tasks and exploit the result of the commonsense constant prediction task to explicitly constrain the constant symbolic table, which can reduce the search space for symbolic generation and ease the difficulty of generating correct constant.   

\noindent\textbf{Self-supervised Number Prediction (SNP) Tasks.} If a solver can fully understand the problem semantics, it should be able to identify the quantity of numbers in a problem (i.e., to count how many numeric values are in the problem) and their corresponding locations in the problem text accurately. For example, if the solver can understand the problem in Fig.~\ref{fig:sns-solver}, it should be able to predict there are two numbers($26$ and $82$) in the problem, and their positions are 15 and 18, respectively. Thus, number quantity prediction and number location prediction are two critical self-supervised tasks to help the problem reader fully understand the problem semantics and measure the ability of problem understanding of a solver. Both two number prediction tasks take the mean of the problem encoder's outputs $\left \{ e_i \right \}^{n}_{i=1}$ as their input and apply a single-layer feed-forward neural network to compute the distribution of number quantity and number locations. The training objectives of two tasks for each problem are formulated as:
\begin{equation}
\begin{aligned}
\mathcal{L}_{NQP} &=-\sum_{i=1}^{Q} qt_i \log p\left(q_i | P \right), \\
\mathcal{L}_{NLP} &=-\sum_{i=1}^{L} lt_i \log p\left(l_{i} | P \right).
\end{aligned}	
\end{equation}
where ${L}_{NQP}$ and ${L}_{NLP}$ denote the loss for the Number Quantity Prediction (NQP) task and Number Location Prediction (NLP) task, respectively. $Q$ and $L$ are the maximum possible quantities of number and maximum possible number locations for a problem at the dataset level. $qt_i$ and $lt_i$ represent the ground-truth value on $i$-th index of the output probability distribution of NQP and NLP, respectively.

\noindent\textbf{Commonsense Constant Prediction (CCP) Task.} Commonsense constants are important for solving some MWPs while most previous methods only consider the constants 1 and 3.14, which are not enough for a solver to solve problems that need other commonsense constants. However, attaching a lot of constants to the problem-specific symbolic table will enlarge the search space, increasing the difficulty of generating rational symbolic equations. Therefore, we propose a commonsense constant prediction task to predict what prior commonsense knowledge (e.g. a chicken has 2.0 legs and a rabbit has 4.0 legs for the problem in Fig.~\ref{fig:sns-solver}) is required for the solver to solve a problem according to the problem context. In this way, we can reduce the search space greatly, thus improving the performance of our solver. Similar to the number prediction tasks, the commonsense constant prediction task takes the mean of the problem encoder's output $\left \{ e_i \right \}^{n}_{i=1}$ as their input and apply a single-layer feed-forward neural network to compute the distribution of number quantity and number locations The training objective for each problem is formulated as:
\begin{equation}
\begin{aligned}
\mathcal{L}_{CCP} & =-\sum_{i=1}^{C} ct_j \log p\left(c_{i} | P \right). \\
\end{aligned}	
\end{equation}
where $C$ is the total number of constants in the symbolic table and $ct_i$ represents the true value on $i$-th index of the output probability distribution. Since it is impossible for the commonsense constant prediction task to achieve 100\% accuracy, in addition to the predicted constants, we add three extra constants that are not predicted but with the highest probability into the symbolic table, making a better trade-off between the size of the search space and prediction accuracy.

\noindent\textbf{Program Consistency Checker (PCC).} Although a problem can be solved by multiple equivalent but different equations, the predicted equations should be consistent with label equations as much as possible in the supervised learning setting. Therefore, we propose a program consistency checker to check the symbolic program consistency and regularize the model by computing semantic loss between the predicted symbolic program and ground-truth equation to ensure the reasonable symbolic equation mapping. Let $\hat{y}_{i}$ and $y_{i}$ represent the predicted symbol and ground-truth symbol, $p_{i}$ represents the probability of $\hat{y}_{i}$, the semantic loss is obtained by computing a distance between the predicted distribution and ground-truth distribution as:
\begin{equation}
\begin{aligned}
\mathcal{L}_{PCC} &= -log\sum_{i}\prod_{\hat{y}_{i}=y_{i}}p_{i}\prod_{\hat{y}_{i}\neq y_{i}}\left ( 1-p_{i}\right ). \\
\end{aligned}
\end{equation}

\noindent\textbf{Duality Exploiting (DE) Task.} Many previous works~\cite{he2016dual,xia2017dual,xiao2018dual,wei2019code} have shown promising results by dual learning framework. Although intuitively, MWP solving and MWP generation are related to each other, i.e., the input of MWP solving is the output of MWP generation, and vice versa, it is very hard for the MWP generation task to generate good enough problems only by the equations without any topic information. Therefore, we propose a duality exploiting task to enhance the understanding ability of our solver by exploiting the quasi duality between symbolic grounded equation generation and the problem's part-of-speech generation. Given a pair of a problem and its corresponding equations ($P$,$T$), and ${P}'$ is the part-of-speech of $P$~\footnote{We use Jieba (https://github.com/fxsjy/jieba) to generate the POS of a problem.}, the training objective of the duality exploiting task is formulated as:
\begin{equation}
\begin{aligned}
\mathcal{L}_{dual}=&\left[\log \hat{p}(P')+\log p\left(T|P\right)- \right. \\
&\left.\log \hat{p}(T)-\log p\left(P'|T\right)\right]^{2}.
\end{aligned}
\end{equation}
where $\hat{p}({P}')$ and $\hat{p}(T)$ are marginal distributions, which can be modeled by their LSTM~\cite{lstm}-based language models, respectively. Besides, we deploy a tree-structure encoder inspired by GTS~\cite{seq2tree} to encode equations in prefix for POS generation.

\subsection{Training Objective}
Given the training dataset $\mathbf{D}$=\{$(P^i,T^1)$, $(P^2,T^2)$, $\cdots$,$(P^N,T^N)$ \}, where $T^i$ is the universal expression tree of problem $P^i$, we minimize the following loss function for our NS-Solver:
\begin{equation}
\begin{aligned}
\mathcal{L} =& \sum_{(P,T)\in \mathbf{D}}\left[\mathcal{L}_{ent1} + \lambda_1 *\mathcal{L}_{dual} + \lambda_2 * \mathcal{L}_{PCC} \right. \\ &\left. + \lambda_3 * \left( \mathcal{L}_{NQP} + \mathcal{L}_{NLP}\right) + \lambda_4 * \mathcal{L}_{CCP} \right].
\end{aligned}
\end{equation}
where
\begin{equation}
\mathcal{L}_{ent1} = -\operatorname{log} \prod_{t=1}^{m}\operatorname{prob}(y_t| P)
\end{equation}
where $m$ denotes the size of T, and $y_t$ denotes the t-th output. $\left \{ \lambda_i \right \}^{4}_{i=1}$ are empirical values that will be detailed in Section~\ref{exp:st}. 

For the duality exploiting task, there is another loss for training the branch of the problem's part-of-speech generation:
\begin{equation}
\mathcal{L}_{POS} = \sum_{(P',T)\in \mathbf{D}}[\mathcal{L}_{ent2} + \lambda_5 *\mathcal{L}_{dual} + \lambda_6 * \mathcal{L}_{PCC'}].
\end{equation}
where
\begin{equation}
\mathcal{L}_{ent2} = -\operatorname{log} \prod_{t=1}^{n}\operatorname{prob}(x_t| T)
\end{equation}
where $n$ denotes the size of P, and $x_t$ denotes the t-th output. $\mathcal{L}_{PCC'}$ is the semantic loss between predicted POS and the ground-truth POS. $\left \{ \lambda_i \right \}^{6}_{i=5}$ are empirical values that will also be detailed in Section~\ref{exp:st}. 
\section{Experiments}

\subsection{CM17K Dataset}
Most public MWPs datasets are quite small such as ALG514 or exist some incorrect labels such as Dolphin18K. An exception is the Math23K dataset, which contains 23161 problems labeled well with structured equations and answers. However, it only contains one-unknown linear math word problems, which is not sufficient to validate the ability of a math solver about solving multiple types of MWPs. Therefore, we introduce a new high-quality math word problems dataset, called CM17K, to validate the universality of a solver and provide a more realistic and challenging benchmark for developing a universal and scalable math solver. We collect CM17K from two education websites\footnote{\href{http://www.zxxk.com/}{http://www.zxxk.com/} and   \href{http://www.jyeoo.com/}{http://www.jyeoo.com/}}. These problems are oriented grades 6-12, containing 4 types of MWPs with more than 17K samples, including 6215 arithmetic MWPs, 5193 one-unknown linear MWPs, 3129 one-unknown non-linear MWPs, and 2498 equation set problems. It should be noticed that our dataset is sufficient for validating the universality of math word problem solvers since these problems can cover most cases about MWPs. We label our data with structured equations and answers following Math23K~\cite{dns}. We split our CM17K into train/valid/test sets at a ratio of 8:1:1. 
\begin{table}[htbp]
\small
\centering
\begin{tabular}{|c|c|c|}
\hline
 & Math23K & CM17K  \\ \hline
\# Avg PL &28.015 &54.365 \\ \hline
\# Avg EL &6.853 &13.853 \\ \hline
\# Avg TS &5.554 &11.834 \\ \hline
\# Avg Num &2.821  &6.383  \\\hline
\# Avg SNI &2.668  &4.111 \\ \hline
\# Avg Ops &3.943  &4.852 \\  \hline
\# Avg Constants & 0.270  & 0.327 \\ \hline
\end{tabular} 
\caption{Statistics of Math23K and CM17K. PL, EL, TS, Num, SNI, Ops, and Constants represent problem length, equation length, equation tree size, number of quantities in problems, number of quantities occurred in both problems and their corresponding equations, number of operators in equations, and number of constants only occurred in equations, respectively.}
\label{tb:data_stat}
\vspace{-3mm}
\end{table}

The data statistics of Math23K and CM17K are shown in Table~\ref{tb:data_stat}. From the statistics, we can see that all statistics of CM17K are larger than Math23K. This shows that our dataset is more challenging and difficult for math word problem solvers. Besides, since CM17K contains more types of MWPs than Math23K, CM17K is more suitable for validating the reasoning ability of a solver than Math23K.  

\subsection{Experimental Setup and Training Details}
\label{exp:st}
\subsubsection{Datasets, Baselines, and Metric} We conduct experiments on Math23K and our CM17K. The main state-of-the-arts to be compared are as follows: \textbf{DNS}~\cite{dns} is a universal solver based on the seq2seq model with significant number identification (SNI). \textbf{GTS}~\cite{seq2tree} is a goal-driven tree-structured MWP solver. \textbf{StackDecoder}~\cite{stackdecoder} is an universal semantically-aligned math word problems solver. ~\cite{tsrmd} is an enhanced GTS with teacher-student distillation and multi-decoder ensemble. Again, following prior works~\cite{dns,stackdecoder,seq2tree}, we use \textit{answer accuracy} as the evaluation metric: if the calculated value of the predicted equation tree equals to the true answer, it is thought as correct since the predicted expression is equivalent to the target expression.

\subsubsection{Implementation Details} We use Pytorch\footnote{http://pytorch.org} to implement our model on Linux with an NVIDIA RTX2080Ti GPU card. All those words with fewer than 5 occurrences are converted into a special token UNK. The size of word embeddings and all hidden states for other layers are set as 128 and 512, respectively. Our model is optimized by ADAM optimizor~\cite{adam} with $\beta_1$ = 0.9, $\beta_2$ =0.999, and $\epsilon$ = $1e^{-8}$. The mini-batch size is set as 32. The initial learning rate is set as $1e^{-3}$ and then decreases to half every 40 epochs. To prevent overfitting, we set dropout rate as 0.5 and weight decay as $1e^{-5}$. Finally, we conduct greedy search to generate symbolic equation trees. We set $\lambda_1$, $\lambda_2$, $\lambda_3$, $\lambda_5$, and $\lambda_6$ as 0.0005, 0.01, 1.0, 0.005, and 0.1 for both datasets, respectively. We set $\lambda_4$ as 0.000001 for Math23K while we set $\lambda_4$ as 1.0 for CM17K. All constants are extracted from the training set. In each epoch, all training data is shuffled randomly and then cut into mini-batches. 

\subsection{Answer Accuracy} 
Following prior works~\cite{dns,stackdecoder,seq2tree}, we conduct 5-fold cross-validation on Math23K. For CM17K, we evaluate the performance on the test set. The results are shown in Table~\ref{tab:all}. From Table~\ref{tab:all}, we can observe that benefiting from the four new auxiliary tasks and neural-symbolic paradigm, our NS-Solver outperforms the baselines on both datasets in terms of answer accuracy. Specifically, for Math23K and CM17K, the accuracy gains of NS-Solver over GTS are 1.37\% and 5.93\%, respectively. Comparing with TSN-MD, our solver outperforms it by about 0.6\% on Math23K. It shows that our model is more feasible for solving multiple types of MWPs. It also shows that our NS-Solver is more effective than other state-of-the-art models on the real-world scenario that needs to solve various MWPs with a unified solver.
\begin{table}[htbp]
\small
\centering
\resizebox{0.8\linewidth}{!}{
\begin{tabular}{|c|c|c|}
\hline
Model & Math23K & CM17K \\ \hline
DNS~\cite{dns} & 58.1\% & 15.93\% \\ \hline
StackDecoder~\cite{stackdecoder} & 66.0\%  & 37.24\% \\ \hline
GTS~\cite{seq2tree} &  74.3\% & 47.12\% \\ \hline
TSN-MD~\cite{tsrmd} &  75.1\% & - \\ \hline
NS-Solver (Ours) & \textbf{75.67\%}& \textbf{54.05\%} \\ \hline
\end{tabular}}
\caption{Model comparison on answer accuracy}
\label{tab:all}
\vspace{-3mm}
\end{table}
\subsection{Comparisons on different subsets} We drill down to analyze the generalization of DNS, GTS, and NS-Solver on different types of MWPs in the test subset of CM17K. Their answer accuracy on different types of MWPs is shown in Table~\ref{tab:subset}. We can observe that our NS-Solver outperforms the other two models by a large margin on all subsets. Specifically, the accuracy gains of our NS-Solver over GTS on four subsets are 3.87\%, 9.12\%, 6.99\%, and 9.44\%. This shows that with the help of four auxiliary tasks, our NS-Solver obtains better generalization ability on multiple types of MWPs than baselines.
\begin{table}[htbp]
\small
\centering
\resizebox{\linewidth}{!}{
\begin{tabular}{|c|c|c|c|c|c|}
\hline
\multicolumn{2}{|c|}{} & arithmetic  & \tabincell{c}{one-unknown \\ linear}  & \tabincell{c}{one-unknown \\ non-linear}  & equation set \\  \hline
\multicolumn{2}{|c|}{Number}  & 619 & 526 & 315  & 244   \\ \hline
\multirow{2}{*}{DNS} & Correct &23 &49 &67 &132   \\ \cline{2-6}
& Accuracy &3.7\% &9.32\%  &21.27\%   & 54.1\% \\ \hline
\multirow{2}{*}{GTS} & Correct &255 &220 &201 &128   \\ \cline{2-6}
& Accuracy  &41.20\% &41.83\%  &63.80\%   & 52.45\% \\ \hline
\multirow{2}{*}{NS-Solver (Ours)} & Correct &279 &268 &223 &151   \\ \cline{2-6}
& Accuracy  &\textbf{45.07\%} &\textbf{50.95\%}  &\textbf{70.79\%}   & \textbf{61.89\%} \\ \hline
\end{tabular}}
\caption{Answer accuracy on CM17K's test subset.} 
\label{tab:subset}
\vspace{-3mm}
\end{table}
\subsection{Performance on Tree Length} 
\begin{figure}[t]
\centering
\includegraphics[width=0.75\linewidth]{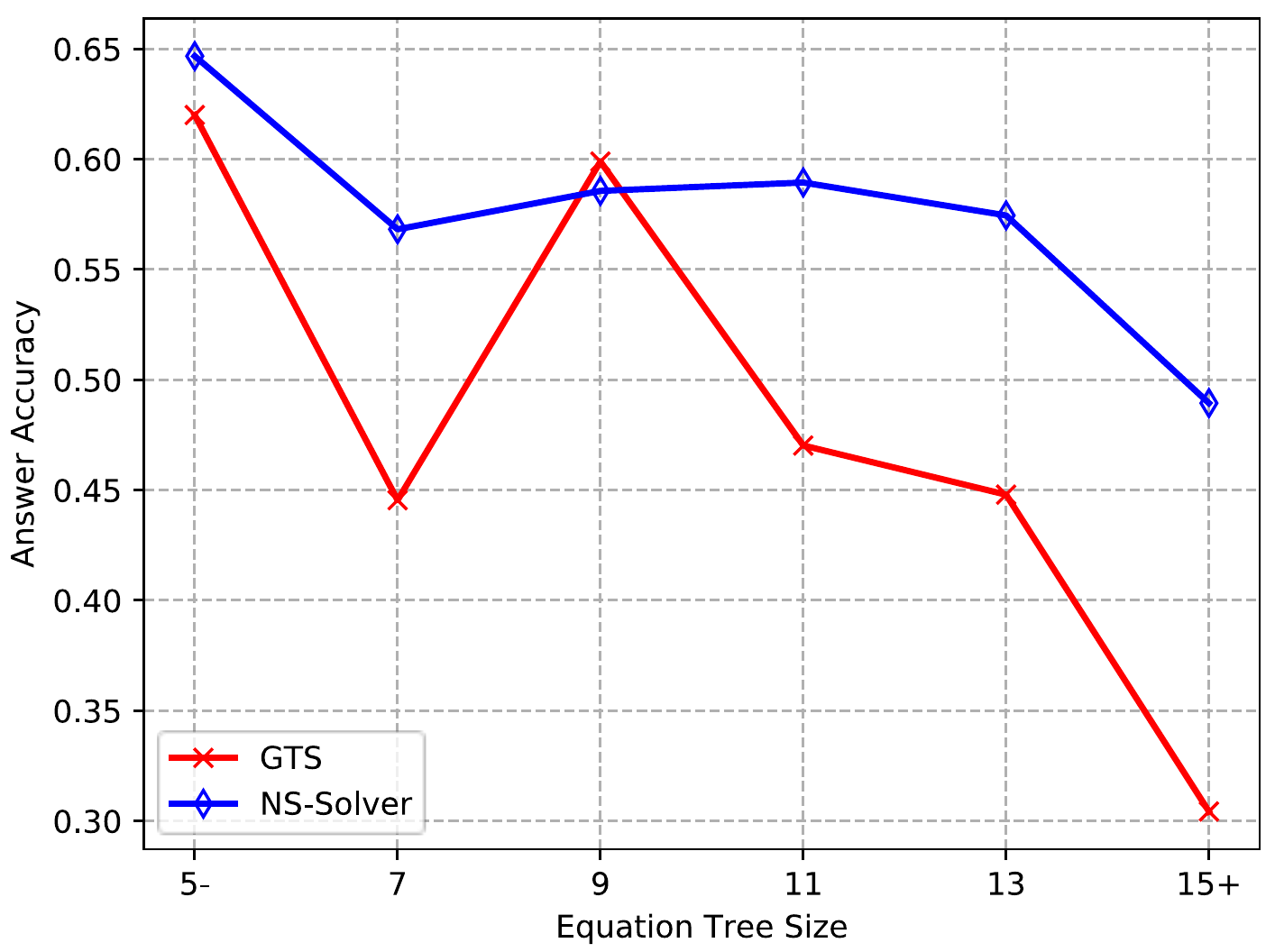}
\caption{Answer accuracies for different sizes of symbolic equation trees on CM17K.} 
\label{tab:error}
\vspace{-3mm}
\end{figure}

Intuitively, the size of the symbolic equation tree is proportional to the complexity of the mathematical relationship in the problem. The more complex the mathematical relationship is, the more difficult it is to solve the problem. Here, we compare our proposed NS-Solver with GTS on CM17K to show the superiority of our NS-Solver on different equation tree sizes. The answer accuracies for different sizes of expression trees on CM17K test subset are shown in Fig.~\ref{tab:error}. We can see that there is a tendency for answer accuracy to degrade with the growth of the problem complexity measured as the size of the equation tree, and our NS-Solver outperforms GTS on most cases of different equation tree sizes. This shows our NS-Solver can better model the mathematical relationships of the problem than GTS. It can also be noticed that the improvement of our NS-Solver over the GTS is increasing when the problems become more complex. 

However, although our model outperforms other methods, there still has room for improvement in semantic understanding and symbolic reasoning since longer equations often match with more complex MWPs which entail more complex math relationships.
\subsubsection{Ablation on different auxiliary tasks} 
\begin{figure}[htbp]
\centering
\includegraphics[width=0.75\linewidth]{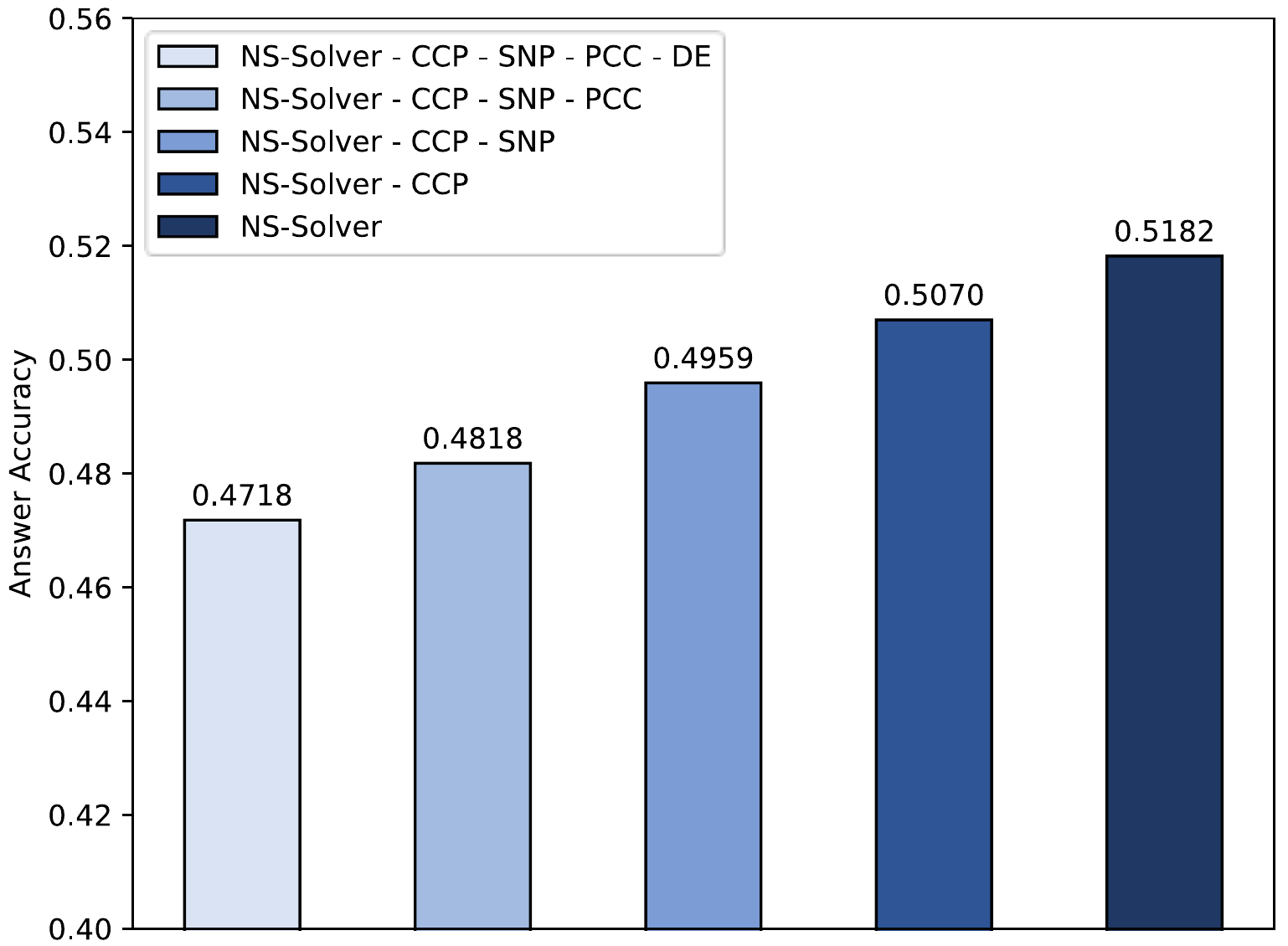}
\caption{Ablation Study on different auxiliary components. ‘-’ represents we remove the component.}
\label{tab:ablation}
\vspace{-3mm}
\end{figure}

We study the contribution of different auxiliary tasks of our NS-Solver. For this purpose, we consider five different combinations: 1) only the backbone [NS-Solver - CCP - SNP - PCC - DE]; 2) backbone + duality exploiting task [NS-Solver - CCP - SNP - PCC]; 3) backbone + duality exploiting task + program consistent checker [NS-Solver - CCP - SNP]; 4) backbone + duality exploiting task + program consistent checker + number prediction tasks [NS-Solver - CCP]; and 5) the proposed NS-Solver [NS-solver]. For each of these combinations, each model was trained for 80 epochs on CM17K and validated on its test subset. The learning rate decreased to half every 20 epochs. The results are provided in Fig.~\ref{tab:ablation}. 

\begin{figure*}[t] 
	\centerline{\includegraphics[width=0.85\textwidth]{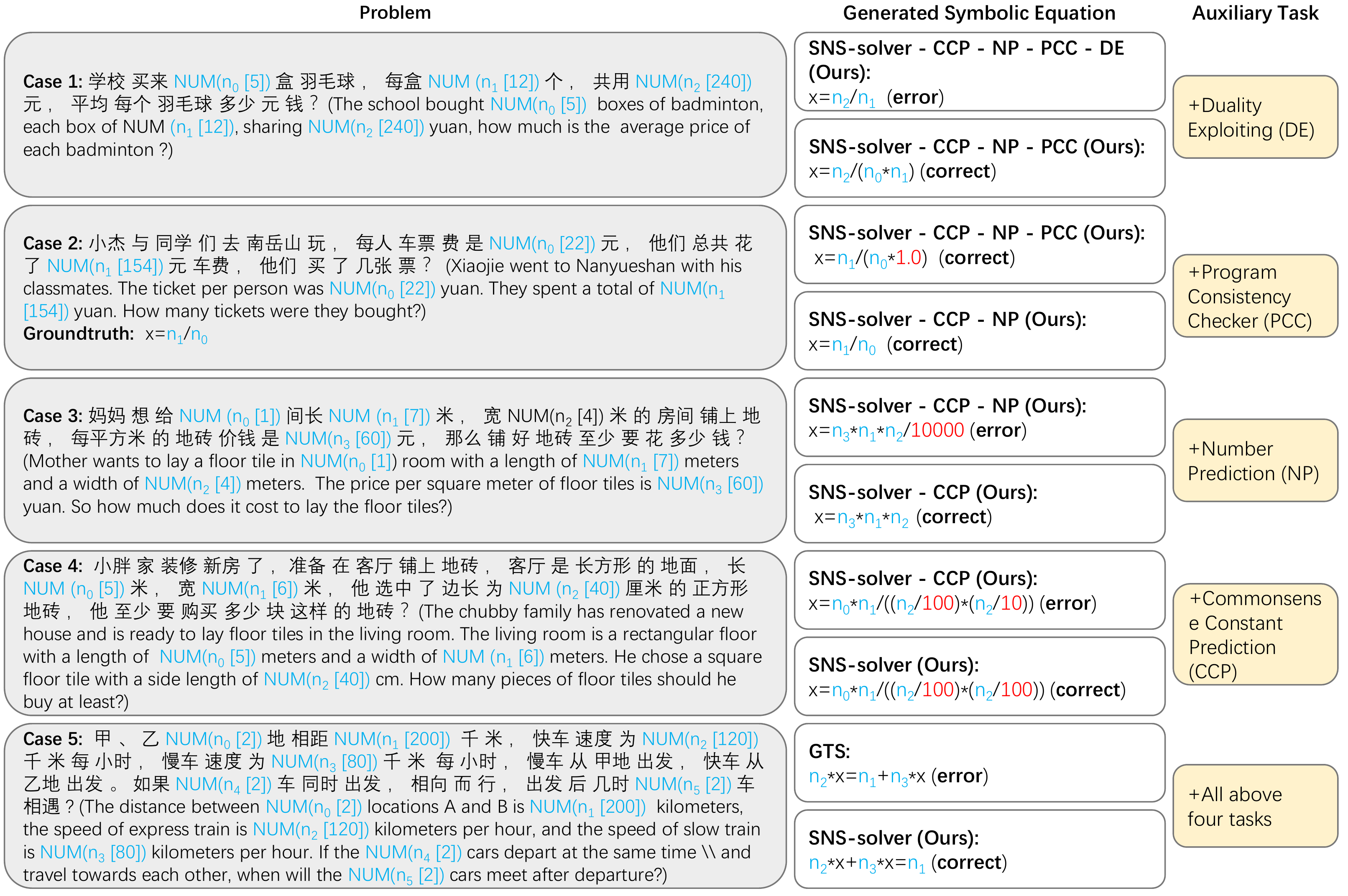}}
	\caption{Typical cases. Note that the results are represented as infix order which is more readable than prefix order. The programs generated by NS-Solver are also translated into human-readable equations. Constants and number symbols are labelled in {\color{red}{red}} and  {\color{cyan}{cyan}}, respectively.}
	\label{tab:cs}
\vspace{-6mm}
\end{figure*}

As one can see, all four auxiliary tasks can improve performance. Specifically, the accuracy gains of DE, PCC, SNP, and CCP are 1.00\%, 1.41\%, 1.11\%, and 1.12\%, respectively. Besides, the binary accuracies of the two SNP tasks are 97\% (number quantity prediction) and 96.8\% (number location prediction). Moreover, the accuracy of our CCP task is 97.8\%. This shows that our auxiliary tasks can enhance our NS-Solver to enforce better problem understanding and symbol reasoning. Overall, our proposed NS-Solver achieves the best answer accuracy.

\subsection{Case Study} We also present the results of our NS-Solver with different combinations of four auxiliary tasks in Fig.~\ref{tab:cs}. Benefiting from explicitly exploiting the probabilistic correlation between two quasi dual tasks to regularize the training process in our duality exploiting (DE) task, our [NS-solver - CCP - SNP - PCC] can generate correct equations by understanding the problem better while [NS-solver - CCP - SNP - PCC - DE] generates error equations, as shown in \textbf{Case 1}. With the program consistency checker (PCC) that effectively regularizes the model's output by constraining the distance between predicted symbols and ground-truth symbols during training, [NS-solver - CCP - SNP] can generate more consistent equations with the ground-truth than [NS-solver - CCP - SNP - PCC], as shown in \textbf{Case 2}. With self-supervised number prediction (SNP), [NS-solver - CCP] can generate better results and avoid generating symbols that do not belong to the problem, as shown in \textbf{Case 3}. With commonsense constant prediction (CCP), our NS-Solver manages to choose correct constants by constraining the constant symbolic table using predicted results of CCP. As shown in \textbf{Case 4}, [NS-solver - CCP] chooses error constant 10 while NS-solver chooses two correct constants. Besides, although GTS and NS-Solver generate the same symbols sometimes, our NS-Solver generates correct equations with the help of our four auxiliary objectives, as shown in \textbf{Case 5}. Overall, all four auxiliary tasks can improve our NS-Solver's understanding and reasoning ability.
\begin{table}[htbp]
\centering
\resizebox{0.99\linewidth}{!}{
\begin{tabular}{|c|c|c|}
\hline
Model & BERT + Tree Decoder~\cite{seq2tree} & NS-Solver + BERT\\ \hline
CM17K & 55.0\% & \textbf{60.68\%} \\ \hline
\end{tabular}}
\caption{Generalization to different backbone}
\label{tab:ext}
\vspace{-3mm}
\end{table}

\subsection{Extends to other backbone}
To show that our auxiliary tasks can be adapted to other backbones, we replace GTS's encoder with BERT (\textbf{BERT + Tree Decoder}) and NS-Solver's encoder with BERT (\textbf{NS-Solver + BERT}), where we adopt a Chinese BERT-base pre-trained with whole word masking~\cite{cui-etal-2020-revisiting}. We conduct experiments on CM17K. The results are shown in Table~\ref{tab:ext}. We can observe that with auxiliary tasks, our \textbf{NS-Solver + BERT} still can outperform \textbf{BERT + Tree Decoder}, which shows that our auxiliary tasks' strong generalization.

\section{Conclusion}
In this work, we propose Neural-Symbolic Solver (NS-Solver) to explicitly and seamlessly incorporate different levels of symbolic constraints by four auxiliary tasks. Our NS-Solver consists of a problem reader to encode problems, a programmer to generate a symbolic grounded program, and a symbolic executor to obtain final results. In addition to supervised learning with target expression, our solver is also optimized via four new auxiliary objectives that enforce four levels of symbolic reasoning. Besides, we also construct a new dataset CM17K containing 4 types of MWPs with more than 17K samples, which provides a more realistic and challenging benchmark for developing a universal and scalable math solver. Extensive experiments on Math23K and  CM17K demonstrate the superiority of our NS-Solver compared to state-of-the-art methods in answer accuracy while ensuring intermediate equation rationality.

\section{Ethical Impact}
We collected CM17K from two online education websites, which is only used for academic research, and the copyright belongs to the original websites. This work may inspire research in the field of numerical reasoning.

\paragraph{Acknowledgements}
This work was supported in part by National Key R\&D Program of China under Grant No.2020AAA0109700, National Natural Science Foundation of China (NSFC) under Grant No.U19A2073, No.61976233 and No. 61836012, the Natural Science Foundation of Guangdong Province under Grant No. 2017A030312006, Guangdong Province Basic and Applied Basic Research (Regional Joint Fund-Key) Grant No.2019B1515120039, Shenzhen Fundamental Research Program (Project No.RCYX20200714114642083 and No.JCYJ20190807154211365), Zhijiang Lab’s Open Fund (No.2020AA3AB14), CSIG Young Fellow Support Fund, and Guangdong Provincial Key Laboratory of Information Security Technology. 

\bibliographystyle{acl_natbib}
\bibliography{acl2021}
\end{document}